\documentclass[11pt]{article}

\usepackage[preprint]{acl}

\usepackage{times}
\usepackage{latexsym}
\usepackage{graphicx}
\usepackage{booktabs}
\usepackage{tabularx}
\usepackage{amsmath}
\usepackage{pifont}
\usepackage{amssymb}
\usepackage{algorithm}
\usepackage{algpseudocode}
\usepackage{booktabs}
\usepackage{multirow}
\usepackage[utf8]{inputenc}
\usepackage{geometry} 
\usepackage{amsmath, amssymb} 
\usepackage{graphicx}         
\usepackage{booktabs}        
\usepackage{hyperref}        
\usepackage[utf8]{inputenc}
\usepackage[most]{tcolorbox} 

\usepackage[T1]{fontenc}

\usepackage[utf8]{inputenc}

\usepackage{microtype}

\usepackage{inconsolata}

\usepackage{graphicx}

\title{CROP: Token-Efficient Reasoning in Large Language Models via Regularized Prompt Optimization}

\author{
 \textbf{Deep Shah\textsuperscript{1}},
 \textbf{Sanket Badhe\textsuperscript{1}},
 \textbf{Nehal Kathrotia\textsuperscript{1}},
 \textbf{Priyanka Tiwari\textsuperscript{2}}
\\
\\
 \textsuperscript{1}Google LLC,
 \textsuperscript{2}Purdue University
}

\newtcolorbox{promptbox}[1][]{
  enhanced,               
  breakable,              
  colback=gray!5!white,
  colframe=gray!75!black,
  fonttitle=\bfseries,
  coltitle=white,
  attach boxed title to top left={yshift=-2mm, xshift=2mm},
  boxed title style={colback=gray!75!black},
  title={Prompt},
  arc=2mm,
  #1
}

\begin{document}
\maketitle
\begin{abstract}
Large Language Models utilizing reasoning techniques improve task performance but incur significant latency and token costs due to verbose generation. Existing automatic prompt optimization(APO) frameworks target task accuracy exclusively at the expense of generating long reasoning traces. We propose Cost-Regularized Optimization of Prompts (CROP), an APO method that introduces regularization on response length by generating textual feedback in addition to standard accuracy feedback. This forces the optimization process to produce prompts that elicit concise responses containing only critical information and reasoning. We evaluate our approach on complex reasoning datasets, specifically GSM8K, LogiQA and BIG-Bench Hard. We achieved an 80.6\% reduction in token consumption while maintaining competitive accuracy, seeing only a nominal decline in performance. This presents a pragmatic solution for deploying token-efficient and cost-effective agentic AI systems in production pipelines.
\end{abstract}

\section{Introduction}
Large Language Models (LLMs) have achieved unprecedented state-of-the-art capabilities in complex reasoning tasks primarily through step-by-step generation strategies like Chain of Thought \cite{wei2022chain,kojima2022large,wang2022self}. By prompting models to decompose intricate problems into intermediate logical steps prior to outputting a final answer, researchers have significantly improved performance across mathematical, logical, and commonsense reasoning benchmarks \cite{nye2021show,zelikman2022star,cobbe2021training}. Reasoning capabilities are an indispensable requirement for deploying foundation models across diverse and critical domains, including financial analysis, algorithmic ranking, medical diagnosis, and legal reasoning \cite{wu2023bloomberggpt, cui2023chatlaw, badhe2026long, shah2026taxonomy}. This paradigm encourages models to mimic structured human deliberation \cite{yao2023tree, besta2024graph}. Consequently, step-by-step rationale generation has become the standard operational mode for extracting high-fidelity reasoning from large-scale foundation models \cite{lightman2024lets}.

Despite these performance gains, the reliance on verbose intermediate reasoning introduces a critical deployment hurdle for real-world production environments. The generation of thousands of intermediate tokens incurs massive inference latency and exorbitant financial costs \cite{pope2022efficiently, aminabadi2022deepspeed, kwon2023efficient}. Recent empirical analyses reveal that modern reasoning models exhibit substantial verbosity compensation, often producing excessively lengthy explanations that provide little additional logical value \cite{patel2024demystify, xu2025chain}. This extensive token consumption thus presents a severe liability that restricts the practical viability of deploying complex reasoning pipelines \cite{li2024benefits, li2024500x, jiang2023llmlingua}.

To optimize these systems, the field has increasingly adopted automated prompt engineering frameworks that leverage LLMs as meta-optimizers \cite{zhou2023large, yang2024large, pryzant2023automatic}. Tools such as DSPy remove the need for manual human trial and error by refining prompts through systematic search \cite{khattab2024dspy}. Recent frameworks extend this paradigm by automating the optimization workflow, utilizing meta-learning and combinatorial search to systematically refine instructions and prompting strategies without human intervention \cite{spiess2025autopdl, murthy2025promptomatix, xu2025metatextgrad}. However, these existing tools suffer from a critical flaw in that they optimize almost exclusively for task accuracy, often at the expense of generating excessively verbose intermediate reasoning traces. By continuously adding edge-case instructions and logic corrections to the prompt, the optimizer inadvertently maximizes the verbosity tax, producing bloated system prompts that force the target model to generate even longer and more exhaustive reasoning traces.

This inherent tendency toward verbosity exposes a fundamental missing mathematical constraint in the current paradigm of textual optimization. In standard machine learning optimization, loss functions routinely incorporate regularization techniques (such as L1 or L2 penalties) to constrain model weights and prevent overly complex architectures from overfitting \cite{goodfellow2016deep, bishop2006pattern}. Yet, textual prompt optimization lacks a true analog to a length penalty.

To solve this structural deficiency, we propose CROP, an applied methodology that introduces an output-biased textual regularizer directly into the prompt optimization loop. CROP builds upon the multi-objective capabilities of textual differentiation to treat output token count as a first-class optimization constraint alongside task accuracy. By computing a continuous length-penalty gradient that functions independently of the task loss, the regularizer explicitly penalizes the target model whenever its output exceeds a minimal threshold. This feedback is then aggregated with traditional accuracy gradients using textual summation. Consequently, CROP forces the meta-optimizer to rewrite system instructions that explicitly constrain verbosity without sacrificing the underlying logical steps necessary for correct inference.

Through this regularized feedback mechanism, CROP successfully forces the discovery of concise reasoning paths. Our empirical evaluations demonstrate that this approach effectively shrinks the output distribution, minimizing both output length and overall inference latency. We evaluate CROP on complex reasoning benchmarks, specifically GSM8K \cite{cobbe2021training}, LogiQA \cite{liu2020logiqa}, and BIG-Bench Hard \cite{suzgun2023challenging, srivastava2023beyond}. The results show that our cost-aware optimization method yields a dramatic reduction in output token consumption by 80.6\% reduction in token consumption while maintaining competitive accuracy. By establishing a continuous textual penalty for verbosity, CROP provides a pragmatic and highly effective solution for deploying token-efficient compound AI systems in strict production pipelines.

\subsection{Main Contributions}
In summary, our primary contributions are as follows:

\begin{itemize}
    \item \textbf{Cost-Aware Prompt Optimization (CROP):} We propose CROP, a novel token-efficient automatic prompt optimization framework. By introducing an response adaptive textual length penalty, CROP successfully decouples reasoning accuracy from generation verbosity during the prompt discovery phase.
    
    \item \textbf{Massive Inference Efficiency:} We empirically demonstrate that CROP reduces output token consumption by up to 80.6\% on complex reasoning benchmarks (GSM8K,  LogiQA and BIG-Bench Hard) while strictly preserving the task accuracy of unconstrained Chain-of-Thought prompting.
    
    \item \textbf{Autonomous Discovery of Symbolic Reasoning:} We show that our regularized textual gradients automatically force the target LLM to adopt highly compressed, symbolic reasoning structures. This emergent behavior conceptually mirrors human-engineered strategies (e.g., ``Chain-of-Draft'') without requiring manual prompt design.
\end{itemize}

\section{Related Work}

\subsection{Automatic Prompt Optimization}

The optimization of natural language prompts has evolved from discrete vocabulary search \cite{shin2020autoprompt, deng2022rlprompt, prasad2023grips, guo2023connecting, chen2024evoprompt, badhe2026prompt} to utilizing large language models as meta-optimizers \cite{zhou2023large, yang2024large, pryzant2023automatic, wang2023instructzero}. Frameworks such as DSPy and TextGrad \cite{khattab2024dspy, yuksekgonul2024textgrad} have formalized this into comprehensive computation graphs, enabling automated textual backpropagation capable of optimizing user-defined metrics and rewards. However, when applied to complex reasoning tasks, textual optimization exposes a fundamental generative bias: because evaluators diagnose logic errors via natural language critiques, they inherently resolve failures by prescribing additive logic and expanded instructions. This accumulation of instructional leads to unchecked increase of reasoning leading to verbosity. Recognizing this computational bottleneck, recent frameworks such as Promptomatix and AutoPDL \cite{pan2024autopdl, murthy2025promptomatix} have introduced cost-aware objectives to balance efficiency with performance, alongside discrete evolutionary algorithms like CAPO and ParetoPrompt \cite{wen2025paretoprompt, mopo2025, zehle2025capo}. Building upon this critical shift, CROP introduces a continuous, dual-objective textual optimization landscape. By applying an unconditional, output-biased regularizer alongside task gradients, CROP exerts continuous downward pressure on token bloat, forcing the meta-optimizer to systematically balance logical correctness with generative brevity in a way that purely additive textual gradients cannot.

\subsection{Inference-Time Scaling and Self-Correction}
Alternative approaches improve reasoning by scaling compute during inference. Methodologies ranging from Self-Consistency \cite{wang2022self} to iterative self-correction frameworks like Reflexion and Self-Refine \cite{shinn2024reflexion, madaan2024selfrefine} allow models to critique and repeatedly refine their intermediate outputs. While yielding substantial accuracy gains, these pipelines inherently multiply the token consumption and latency of the system with every additional trial. In contrast to these methods that severely compound inference overhead, CROP shifts the iterative refinement burden entirely to an offline prompt optimization phase, ultimately producing a single, highly efficient system prompt that elicits compressed reasoning trajectories in a single forward pass.

\subsection{Cost-Aware Systems and Verbosity Control}
To mitigate computational overhead \cite{pope2022efficiently, kwon2023efficient}, researchers have developed system-level routing protocols \cite{chen2023frugalgpt, ong2024routellm, aggarwal2024automix, jian2023ecoassistant}, and inference-time early exiting or truncation strategies \cite{ mao2025early, jiang2023llmlingua, wu2024empirical}. Furthermore, recent literature focuses on explicitly compressing reasoning traces via continuous latent projections \cite{hao2024coconut, shen2025codi, wei2025simcot}, conditioned textual pruning \cite{kang2024c3ot, xia2025tokenskip, wang2025compressing}. However, these approaches introduce severe operational trade-offs. Methods relying on continuous latent projections or explicit fine-tuning are computationally prohibitive and alter foundation weights. Conversely, structured static prompting remains notoriously brittle, while dynamic token-pruning and step-skipping often require specialized decoding mechanisms or architectural interventions. Unlike approaches that demand expensive weight updates, custom decoding algorithms, or fragile hardcoded constraints, CROP operates entirely in the natural language optimization space. It dynamically modulates verbosity at the prompt level, systematically discovering optimal, token-efficient reasoning pathways through regularized textual feedback.

\begin{figure*}
    \centering
    \includegraphics[width=0.95\textwidth]{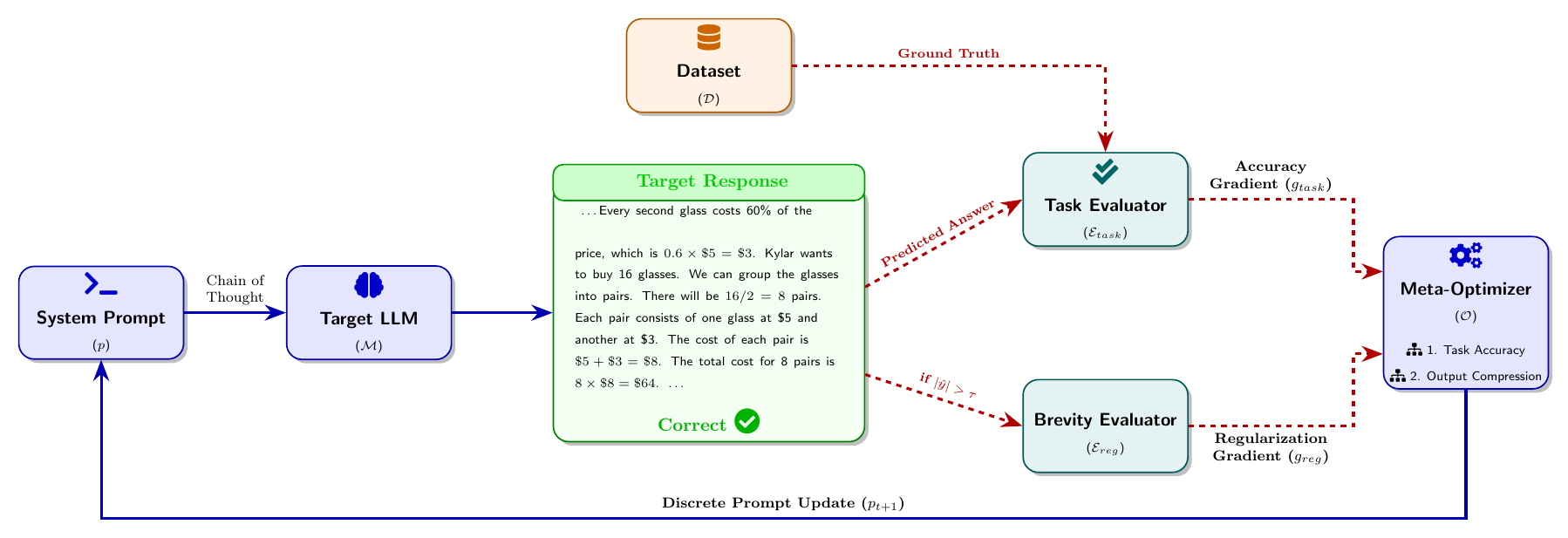}
    \caption{\textbf{Overview of CROP optimization pipeline.} A meta-optimizer updates the system prompt by aggregating textual gradients for both task accuracy ($g_{task}$) and response brevity ($g_{reg}$), systematically enforcing token-efficient reasoning.}
    \label{fig:framework}
\end{figure*}

\section{Methodology}
\label{sec:methodology}

In this section, we formalize \textbf{CROP}, a framework that extends the automatic prompt optimization (APO) paradigm to resource-constrained environments. We first briefly review the preliminary formulation of textual differentiation, and subsequently detail our novel textual regularization objective.

\subsection{Preliminaries: Textual Differentiation}
Let $\mathcal{V}^*$ denote the space of all possible natural language strings. In a standard LLM-based computation graph, we define a system prompt $p \in \mathcal{V}^*$ as a learnable parameter, an input query $x \in \mathcal{V}^*$, and a target model $\mathcal{M}$. The forward pass generates an output $\hat{y} = \mathcal{M}(x; p)$. 

Given a ground-truth label $y$, a textual loss function $\mathcal{L}_{\text{task}}$ evaluates the prediction. Instead of yielding a scalar gradient, textual backpropagation queries a secondary evaluator LLM  $\mathcal{E}$, to generate a \textit{textual gradient} $g_{\text{task}} \in \mathcal{V}^*$. This gradient is a natural language critique detailing the logical flaws in $\hat{y}$ and suggesting semantic updates to $p$:
\begin{equation}
    g_{\text{task}} = \mathcal{E}_{\text{task}}(p, x, \hat{y}, y)
\end{equation}
An optimizer LLM, $\mathcal{O}$, then acts as the parameter update function, synthesizing the batch of gradients to perform a discrete step in the prompt space: $p_{t+1} = \mathcal{O}(p_t, \{g_{\text{task}}^{(i)}\}_{i=1}^B)$, where $B$ is the batch size.

\subsection{Verbose Reasoning}
While $\mathcal{L}_{\text{task}}$ effectively maximizes reasoning accuracy, it is agnostic to computational efficiency. Consequently, the meta-optimizer $\mathcal{O}$ inherently gravitates toward system prompts that trigger useful but overly verbose Chain-of-Thought (CoT) bloat in $\hat{y}$. In production systems, the length of $\hat{y}$ directly influences the inference latency and compute requirements. 


\begin{algorithm}[!ht]
\caption{CROP: Cost-Regularized Optimization of Prompts}
\label{alg:reggrad}
\textbf{Require:} Initial system prompt $p_0$, Target LLM $\mathcal{M}$, Meta-Optimizer LLM $\mathcal{O}$ \\
\textbf{Require:} Task Evaluator $\mathcal{E}_{\text{task}}$, Brevity Evaluator $\mathcal{E}_{\text{reg}}$ \\
\textbf{Require:} Dataset $\mathcal{D}$, Batch size $B$, Max iterations $T$
\begin{algorithmic}[1]
\State $p \gets p_0$
\For{$t = 1$ \textbf{to} $T$}
    \State Sample batch $\{(x_i, y_i)\}_{i=1}^B \sim \mathcal{D}$
    \State $G_{\text{batch}} \gets \emptyset$ \Comment{Initialize empty batch gradient list}
    
    \For{$i = 1$ \textbf{to} $B$}
        \State $\hat{y}_i \gets \mathcal{M}(x_i; p)$ \Comment{Forward Pass}
        
        \State $g_{\text{task}}^{(i)} \gets \mathcal{E}_{\text{task}}(p, x_i, \hat{y}_i, y_i)$ \Comment{Compute Task Accuracy Gradient}
        
        \State $g_{\text{reg}}^{(i)} \gets \mathcal{E}_{\text{reg}}(p, \hat{y}_i)$ \Comment{Compute Textual Regularization Gradient}
        
        \State $g_{\text{total}}^{(i)} \gets g_{\text{task}}^{(i)} \oplus \text{``\textbackslash n''} \oplus g_{\text{reg}}^{(i)}$ \Comment{Aggregate Gradients via Concatenation}
        \State $G_{\text{batch}} \gets G_{\text{batch}} \cup \{g_{\text{total}}^{(i)}\}$
    \EndFor
    
    \State $\text{Accuracy}_t \gets C_{\text{correct}} / B$
    \State $\text{AvgLength}_t \gets L_{\text{total}} / B$
    \State $S_t \gets \text{Accuracy}_t - \lambda \cdot L_{\text{norm}}(t)$ \Comment{Evaluate trade-off score}
    
    \If{$S_t > S_{\text{best}}$} \Comment{Update best prompt if score improves}
        \State $S_{\text{best}} \gets S_t$
        \State $p^* \gets p$
    \EndIf
    
    \State $p \gets \mathcal{O}(p, G_{\text{batch}})$ \Comment{Meta-Optimizer computes discrete prompt update}
\EndFor
\State \textbf{return} $p^*$ \Comment{Optimized token-efficient prompt}
\end{algorithmic}
\end{algorithm}

\subsection{Textual Regularization}
CROP applies $\mathcal{L}_{\text{reg}}$ across the entire batch, asserting a constant downward pressure on token consumption. We define a secondary evaluator prompt, $\mathcal{E}_{\text{reg}}$, which receives the generated output and computes a textual brevity gradient:
\begin{equation}
    g_{\text{reg}} = \mathcal{E}_{\text{reg}}(p, \hat{y})
\end{equation}

The resulting gradient $g_{\text{reg}}$ explicitly instructs the optimizer to inject brevity constraints into $p$. The regularization gradient is computed only when $\|\hat{y}\|$ exceeds a predefined threshold, preventing the penalization of already token-efficient reasoning. To maintain optimization stability and prevent destructively large steps in the prompt space—which frequently induce catastrophic forgetting of the underlying logic—$\mathcal{E}_{\text{reg}}$ is constrained to enforce only a gradual, incremental reduction in output length per iteration. Furthermore, the evaluator is explicitly directed to target peripheral narrative elements and conversational filler, strictly preserving the tokens associated with critical computational steps.

\subsection{Gradient Aggregation and Multi-Objective Optimize}
To perform the parameter update, CROP treats prompt optimization as a formal multi-objective problem. We leverage the concatenation properties of textual autograd engines to compute the total gradient. For a single instance, the combined textual gradient $g_{\text{total}}$ is defined via the string concatenation operator $\oplus$:
\begin{equation}
    g_{\text{total}} = g_{\text{task}} \oplus \text{``\textbackslash n''} \oplus g_{\text{reg}}
\end{equation}

During the batched optimization step, the meta-optimizer $\mathcal{O}$ receives the aggregated sets of both task gradients and regularization gradients and performs the update step $p_{t+1} = \mathcal{O}(p_t, \mathbf{g}_{\text{total}})$. 

Rather than naively returning the prompt from the final iteration, CROP employs a structured selection criterion to identify the optimal prompt $p^*$ discovered during the optimization trajectory. We define a composite score $S$ for each candidate prompt:
\begin{equation}
    S(p) = \text{Accuracy}(p) - \lambda \cdot L_{\text{norm}}(p)
\end{equation}
where $L_{\text{norm}}(p)$ represents the generated token count normalized by the average COT output token length, and $\lambda$ is a tunable regularization coefficient. The magnitude of $\lambda$ dictates the explicit trade-off between task performance and token consumption. We selected $\lambda$ through an empirical grid search on the validation set over the range of $0.01$ to $0.5$. We identified $\lambda = 0.2$ as the optimal configuration, as it provided the maximum reduction in reasoning verbosity without inducing a statistically significant degradation in task accuracy.

\section{Experiment Setup}

\subsection{Model Setup}
\label{subsec:model_setup}
We assign specific Large Language Models to the distinct roles of evaluation, optimization, and inference. We utilize Gemini 2.0 Flash as the dual-evaluator ($\mathcal{E}_{\text{task}}$ and $\mathcal{E}_{\text{reg}}$) to compute task accuracy critiques and regularized verbosity gradients. For the parameter update step, we deploy Gemini 3.1 Pro (with Thinking enabled) as the meta-optimizer ($\mathcal{O}$). A high-capacity model is critical here, as synthesizing multi-objective, conflicting textual gradients requires an extended context window and advanced deductive reasoning (ablated in Section~\ref{sec:results}).

For forward-pass inference ($\mathcal{M}$), we evaluated two models: Gemini 2.0 Flash and Qwen 2.5 7B.

\subsection{Baselines}
\label{subsec:baselines}

We benchmark CROP against zero-shot prompting, chain-of-thoughts, concise induced prompting and automatic prompt optimization method TextGrad. We intentionally omit static length-constrained prompts, which are brittle and degrade reasoning; furthermore, the model was unable to follow explicit token-limit instructions and frequently exceeded the given limits.

\begin{itemize}
    \item \textbf{Standard Zero-Shot (Direct Prompting):} Prohibits intermediate reasoning steps to output the final answer directly. This establishes our empirical lower bound for token consumption, though typically at the severe expense of task accuracy.
    
    \item \textbf{Chain-of-Thought (CoT):} Appends a standard step-by-step reasoning directive. This serves as our upper bound for baseline accuracy, simultaneously representing the widely used but verbosity prompting technique.

    \item \textbf{Conciseness inducing prompts:} We used different prompts which encourages concise reasoning via prompting \cite{lee2025well}. (See prompts in Appendix \ref{baseline_prompts})

    \item \textbf{TextGrad:} An automatic prompt optimization framework that utilizes textual gradients to maximize task accuracy without penalizing verbose reasoning during the optimization loop.
\end{itemize}

\begin{figure*}
    \centering
    \includegraphics[width=0.8\linewidth]{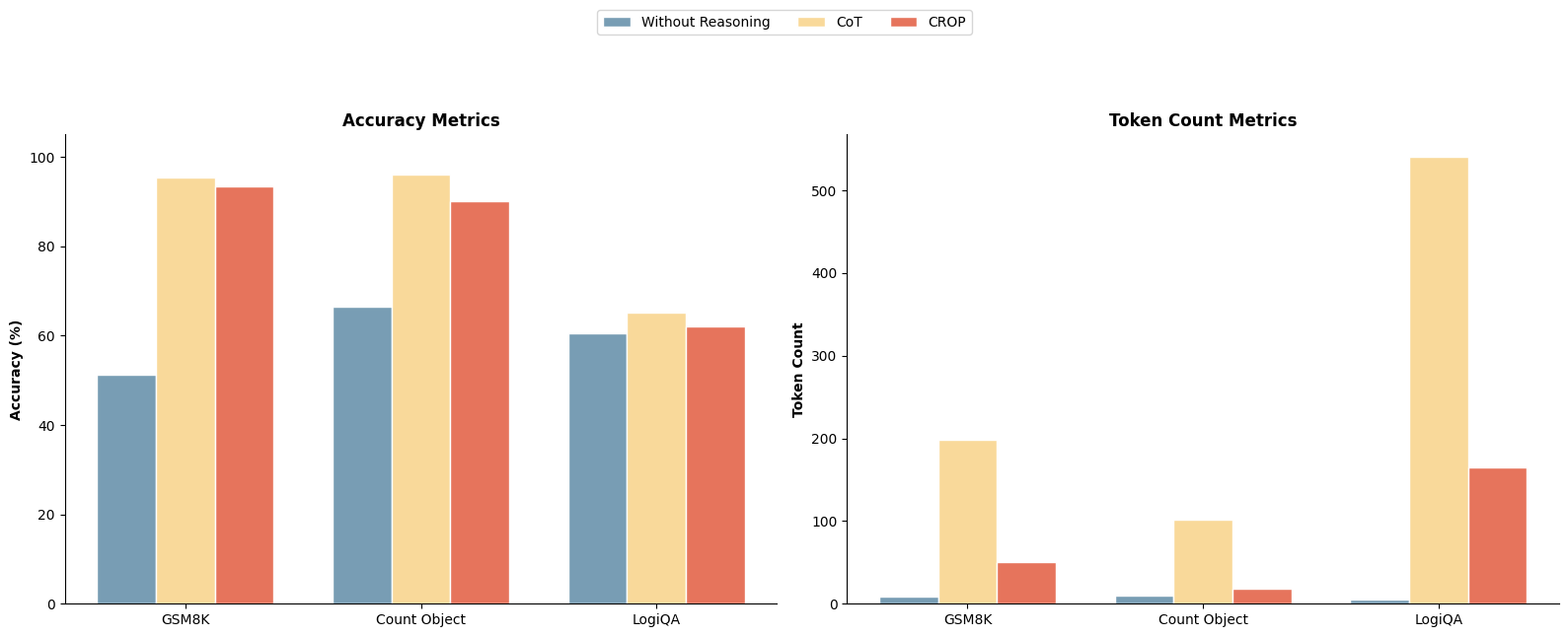}
    \caption{Performance and efficiency comparison across three datasets using the Gemini 2.0 Flash model. }
    \label{fig:placeholder}
\end{figure*}

\subsection{Datasets}
\label{subsec:datasets}

To evaluate CROP across diverse cognitive domains, we selected three challenging reasoning benchmarks. Because these tasks inherently necessitate intermediate logic, they intrinsically incentivize unconstrained LLMs to generate highly verbose Chain-of-Thought (CoT) traces, providing an optimal testbed for our compression framework. To prevent data leakage, we utilize the official train, validation, and test splits provided by the dataset authors wherever available. For the BBH Object Counting task, which lacks standard splits, we manually partition the data into a 25/25/50 split for training, validation, and testing, respectively.

\begin{itemize}
    \item \textbf{GSM8K \cite{cobbe2021training}:} A benchmark of linguistically diverse grade-school math word problems. This dataset requires accurate multi-step arithmetic reasoning, which frequently induces unconstrained LLMs to produce excessively long textual explanations for basic algebraic steps.
    
    \item \textbf{Object Counting (BIG-Bench Hard) \cite{suzgun2023challenging}:} An algorithmic reasoning task requiring the model to determine the total count of a specific semantic class from a natural-language collection of items. This tests symbolic tracking, where standard CoT prompts typically trigger exhaustive, token-heavy item enumeration.
    
    \item \textbf{LogiQA \cite{liu2020logiqa}:} A challenging machine reading comprehension dataset sourced from expert logical examinations. It requires complex deductive inference (e.g., categorical, conditional, and disjunctive reasoning), assessing CROP's ability to maintain the structural integrity of formal logical arguments under aggressive token compression.
\end{itemize}

\begin{table}[ht]
\centering
\begin{tabularx}{\linewidth}{>{\bfseries\hsize=0.3\hsize}X >{\hsize=0.7\hsize}X}
\toprule
Method & Core Instruction \\
\midrule
No Reasoning & Output ONLY the final answer without reasoning or thinking. \\
\midrule
Chain-of-Thought & Think step by step. \\
\midrule
CROP (Ours) & [\dots] concisely, showing the combined essential calculations leading directly to the final numerical answer. Omit intermediate labels. Focus on chained calculations. \\
\bottomrule
\end{tabularx}
\caption{Comparison of core system instructions across prompting baselines for the GSM8K dataset. Standard prefix formulations and output formatting constraints have been omitted to highlight the distinct reasoning directives.}
\label{tab:prompt_comparison}
\end{table}

\begin{table}[ht]
\centering
\begin{tcolorbox}[colback=blue!2, colframe=black!75, title=Qualitative comparison of generated reasoning traces on the GSM8K dataset., fonttitle=\bfseries\small, arc=1mm, boxrule=0.5pt, left=4pt, right=4pt, top=4pt, bottom=4pt]
\small
\textbf{Input Query:} Kylar went to the store to buy glasses for his new apartment. One glass costs \$5, but every second glass costs only 60\% of the price. Kylar wants to buy 16 glasses. How much does he need to pay for them?\\

\vspace{1mm}
\hrule
\vspace{2mm}

\textbf{1. Direct Prompting (No Reasoning)} \\
68 \\
\vspace{-2mm}
\begin{flushright}
\textcolor{red}{\textbf{\ding{55} Incorrect}}
\end{flushright}

\vspace{-1mm}
\hrule
\vspace{2mm}

\textbf{2. Chain-of-Thought (Standard Verbosity)} \\
Let the cost of one glass be \$5. Every second glass costs 60\% of the price, which is 0.6 * \$5 = \$3. Kylar wants to buy 16 glasses. We can group the glasses into pairs. There will be 16 / 2 = 8 pairs. Each pair consists of one glass at \$5 and another at \$3. The cost of each pair is \$5 + \$3 = \$8. The total cost for 8 pairs is 8 * \$8 = \$64.\\
\vspace{-2mm}
\begin{flushright}
\textcolor{green!70!black}{\textbf{\ding{51} Correct (128 tokens)}}
\end{flushright}

\vspace{-1mm}
\hrule
\vspace{2mm}

\textbf{3. CROP (Ours Compressed Reasoning)} \\
8 * 5 + 8 * (5 * 0.6) = 40 + 8 * 3 = 40 + 24 = 64 \\
\vspace{-2mm}
\begin{flushright}
\textcolor{green!70!black}{\textbf{\ding{51} Correct (44 tokens)}}
\end{flushright}

\end{tcolorbox}
\caption{Reasoning comparison on the GSM8K dataset.}
\label{tab:gsm8k_glasses}
\end{table}



\begin{table*}[t]
\centering
\caption{Model Performance Across Prompting Methods. Cells display Accuracy (\%) with Average Token Count in parentheses.}
\label{tab:combined_performance}
\resizebox{\textwidth}{!}{%
\begin{tabular}{llccccccc}
\toprule
\textbf{Model} & \textbf{Dataset} & \textbf{CoT} & \textbf{w/o Reasoning} & \textbf{CROP} & \textbf{TextGrad} & \textbf{Be Concise} & \textbf{Only Number} & \textbf{Use Abbrev} \\
\midrule
\multirow{3}{*}{\textbf{Gemini 2 Flash}} 
& GSM8K        & 95.3 (198.4) & 51.2 (9.2) & 93.4 (50.0)  & 95.5 (185.5) & 96.2 (120.8) & 91.4 (113.0) & 95.1 (145.7) \\
& Count Object & 96.0 (101.2) & 66.4 (9.4) & 94.8 (19.6)  & 90.5 (24.5)  & 96.0 (59.0)  & 92.5 (55.0)  & 95.0 (92.0)  \\
& LogiQA       & 65.2 (540.5) & 60.4 (5.1) & 64.2 (181.0) & 66.4 (663.0) & 65.1 (301.0) & 61.9 (423.0) & 65.3 (422.0) \\
\midrule
\multirow{3}{*}{\textbf{Qwen2.5 7B}}     
& GSM8K        & 90.2 (123.4) & 54.9 (3.4) & 87.8 (21.3)  & 89.8 (101.2) & 90.4 (82.3)  & 85.1 (31.1)  & 88.2 (72.4)  \\
& Count Object & 92.1 (56.3)  & 67.8 (3.8) & 90.1 (12.1)  & 88.8 (32.7)  & 92.7 (46.6)  & 81.5 (16.2)  & 90.2 (39.7)  \\
& LogiQA       & 61.2 (287.0) & 54.0 (3.4) & 59.3 (84.2)  & 60.1 (212.3) & 61.1 (152.3) & 50.8 (32.8)  & 54.7 (212.0) \\
\bottomrule
\end{tabular}%
}
\end{table*}

\section{Results and Analysis}
\label{sec:results}

We evaluate CROP's capacity to eliminate the computational redundancy inherent in unconstrained Chain-of-Thought (CoT) reasoning. By integrating a continuous brevity gradient into the optimization loop, CROP explicitly optimizes for reasoning density alongside task performance. As detailed in Table~\ref{tab:combined_performance}, our framework compresses output token consumption by up to 80.6\% without degrading exact-match accuracy. These results establish that high-fidelity inference can be sustained at a fraction of the standard computational cost.

\subsection{Significant Reductions in Token Consumption}
CROP consistently achieved state-of-the-art token efficiency while preserving the logical integrity required for complex problem-solving. On mathematical arithmetic (GSM8K), algorithmic tasks (Object Counting), and LogiQA, the framework reduced output token consumption by 74.8\%, 80.6\%, and 66.5\% respectively while retaining reasoning leading to comparable accuracy for Gemini 2.0 Flash. Similar token-efficiency was observed for Qwen 2.5 7B as well. This starkly contrasts with the Direct Prompting (Without Reasoning) baseline, which collapses entirely on these tasks. This divergence confirms that while intermediate computation is strictly necessary for accuracy, the narrative \textit{verbosity} of that computation is highly compressible via regularized feedback.

\subsection{Emergent Symbolic Reasoning}
A qualitative examination of the optimized prompts (Table \ref{tab:prompt_comparison}) and generated outputs (Table \ref{tab:gsm8k_glasses}) reveals that CROP discovered prompts fundamentally shifts how the target model structures its intermediate logic. Rather than generating conversational heuristics (e.g., ``First, let us observe that...''), the regularized prompts elicit a dense, symbolic syntax. On GSM8K, the model spontaneously adopts a mathematical shorthand analogous to the recently proposed ``Chain-of-Draft'' \cite{xu2025chain} methodology. However, whereas prior works rely on manual human engineering to elicit this behavior, CROP discovers this optimal reasoning syntax entirely autonomously. 

\begin{table}[ht]
\centering
\begin{tabularx}{\linewidth}{>{\bfseries\hsize=0.28\hsize}X >{\hsize=0.72\hsize}X}
\toprule
Meta-Optimizer Model & Discovered System Prompt (Core Instructions) \\
\midrule
Gemini 2.0 Flash \newline \textit{(Lower Capacity)} & Solve the problem. Only provide the answer and do not provide any explanations. \\
\midrule
Gemini 3.1 Pro \newline \textit{(High Capacity)} & Solve the problem with extreme conciseness. Do not use conversational filler, restate the question, or write full sentences. Simply provide the final sum. \\
\bottomrule
\end{tabularx}
\caption{Comparison of optimized system prompts generated by different meta-optimizers for the BBH Object Counting task. While the lower-capacity model (Gemini 2.0 Flash) defaults to a blunt truncation directive, the high-capacity model (Gemini 3.1 Pro) successfully synthesizes multi-faceted constraints targeting specific vectors of verbosity.}
\label{tab:optimizer_capacity}
\end{table}

\begin{table}[ht]
\centering
\begin{tabularx}{\linewidth}{>{\bfseries\hsize=0.35\hsize}X >{\hsize=0.65\hsize}X}
\toprule
Optimization Setting & Discovered System Prompt (Core Instructions) \\
\midrule
Low Capacity Optimizer \newline \textmd{(Gemini 2.0 Flash, $B=128$)} & Solve the problem. Only provide the answer and do not provide any explanations. \\
\midrule
Stochastic Optimization \newline \textmd{(Gemini 3.1 Pro, $B=1$)} & Solve the problem concisely. Go straight to the answer without restating the problem or adding redundant explanatory sentences. \\
\midrule
Optimal Configuration \newline \textmd{(Gemini 3.1 Pro, $B=128$)} & Solve the problem with extreme conciseness. Do not use conversational filler, restate the question, or write full sentences. Simply provide the final sum. \\
\bottomrule
\end{tabularx}
\caption{Qualitative ablation of discovered system prompts under different optimization constraints for the BBH dataset. Only the optimal configuration (high capacity, large batch) successfully balances extreme conciseness with the preservation of reasoning.}
\label{tab:ablation_prompts}
\end{table}

\subsection{Role of Optimizer Scale in Generalization}
The meta-optimizer ($\mathcal{O}$) is responsible for synthesizing batched textual accuracy and regularization gradients into a coherent, updated system prompt. In early iterations, deploying a lower-capacity model (Gemini 2.0 Flash) as the meta-optimizer resulted in a severe degradation of prompt quality. The optimizer consistently overfit to idiosyncratic errors from isolated batch examples rather than deriving generalizable instructional updates. Extensive tuning of the meta-optimizer's system prompt failed to mitigate this limitation. 

Ultimately, scaling the meta-optimizer to a frontier reasoning model—specifically Gemini 3.1 Pro with thinking capabilities enabled—resolved these issues and produced highly robust system prompts (See table \ref{tab:optimizer_capacity}). We hypothesize that synthesizing complex, multi-objective textual feedback requires both the extended context window and the advanced deductive reasoning capabilities inherent to this larger model. 

\subsection{Gradient Stability and Batch Dynamics}
In standard deep learning, gradient variance is inversely proportional to batch size. We observe an analogous phenomenon in regularized textual differentiation. As demonstrated in Table~\ref{tab:ablation_prompts}, ablation experiments utilizing stochastic updates (batch size of 1) yield extreme variance in the textual gradients. When evaluating an isolated example, the meta-optimizer frequently overfits to the length penalty, prioritizing output compression over task accuracy and aggressively stripping away critical cognitive instructions. This results in prompt collapse: the generation of degenerate directives that entirely eliminate intermediate reasoning capacity. Conversely, scaling to a batch size of 128 stabilizes the multi-objective feedback. By providing the meta-optimizer with a statistically representative sample of both reasoning failures and excess verbosity, large batches prevent regularization-induced collapse and ensure stable, convergent trajectories during prompt optimization.

\section{Conclusion}
The deployment of Large Language Models in complex reasoning pipelines has historically been bottlenecked by the severe latency and financial costs associated with verbose Chain-of-Thought generation. In this work, we introduced CROP, a novel multi-objective automatic prompt optimization framework. By integrating a continuous textual regularization penalty directly into the prompt optimization loop, CROP forces the meta-optimizer to navigate the Pareto frontier of accuracy and token-consumption.

Our empirical evaluations demonstrate that intermediate reasoning is highly compressible. CROP successfully reduces output token consumption by up to 80.6\% on rigorous reasoning benchmarks (GSM8K, BIG-Bench Hard, and LogiQA) while maintaining comparable accuracy compared to unconstrained CoT and automatic prompt optimization baselines. Furthermore, our analysis reveals that this dual-objective optimization automatically elicits highly efficient, symbolic reasoning structures, conceptually mirroring manual ``Chain-of-Draft'' prompting without requiring human intervention. Through systematic ablations, we also identified that successful multi-objective textual optimization requires frontier-level model capacity and large batch sizes to stabilize gradient variance and prevent the collapse of intermediate logic.

Ultimately, CROP provides a highly practical, deployment-ready methodology. By systematically excising computational redundancy at the prompt level rather than altering foundational weights, it enables the scalable and cost-effective integration of high-fidelity reasoning models into strict, real-world production and agentic environments.




\section*{Limitations}
While CROP significantly reduces inference-time overhead, we acknowledge two primary limitations regarding its current operational scope. 

One limitation of our current evaluation is that we do not explicitly consider input token length or cost. In modern production environments, prompt caching (or prefix caching) is a default feature that automatically stores and reuses the initial segments of instructions. Because our framework produces a single, static system prompt that remains identical across all user queries, it perfectly leverages this caching mechanism. By reusing this prior computational work, prefix caching reduces time-to-first-token latency by up to 80\% and decreases input token costs by up to 90\% \cite{openai_prompt_caching}. Consequently, the financial impact of a slightly longer optimized system prompt is effectively marginalized. This justifies our exclusive focus on constraining the verbosity of the generated output, which remains the primary driver of unmitigated inference costs.

Second, the optimization phase of CROP necessitates access to high-capacity reasoning models (e.g., Gemini 3.1 Pro with Thinking). We observe that such model is required to effectively synthesize and balance conflicting objectives between task accuracy and textual regularization. Consequently, the efficacy of using smaller, mid-tier models as meta-optimizers for this dual-objective task has not yet been established.

\bibliography{custom}

\onecolumn
\appendix

\section{Example Appendix}
\label{sec:appendix}

\subsection{Baseline prompts}
\label{baseline_prompts}

\begin{promptbox}[title=CoT]
Think Step by Step. 
\end{promptbox}

\begin{promptbox}[title=Be concise]
Think Step by Step. \textbf{Be Concise.}
\end{promptbox}

\begin{promptbox}[title=Only Num/Eq]
Think Step by Step. \textbf{Only use numbers or equations.}
\end{promptbox}

\begin{promptbox}[title=AbbrevWords]
Think Step by Step. \textbf{Abbreviate words as much as possible.}
\end{promptbox}

\begin{promptbox}[title=UseOnlyNTokens]
Think Step by Step. \textbf{Do not use more than N words.}
\end{promptbox}

\subsection{Initial prompts}

\begin{promptbox}[title=GSM8K]
    You will answer a mathemetical reasoning question.
    
The last line of your response should be of the following format: 'Answer: \$VALUE' where VALUE is a numerical value.
\end{promptbox}

\begin{promptbox}[title=BBH Object counting]
    Solve the problem
    
The last line of your response should be of the following format: 'Answer: VALUE' where VALUE is a numerical value.
\end{promptbox}

\begin{promptbox}[title=LogiQA]
    You will answer a logical reasoning question.
    
The last line of your response should be of the following format: 'Answer: VALUE' where VALUE is the zero-based numerical index of the correct answer.
\end{promptbox}

\subsection{TextGrad Learnt Prompts}

\begin{promptbox}[title=GSM8K]
    You will answer a mathematical reasoning question.
First, explicitly and concisely state the final goal of the question.

Think step by step and be extremely concise in your reasoning. Avoid conversational filler and jump straight into the math.

Define your variables clearly and explicitly track units during your intermediate calculations to prevent conversion errors. 

Carefully consider edge cases, hidden assumptions, overlapping sets, and ensure your final unit matches what the question asks for (e.g., dollars vs. cents). Double-check your arithmetic. 

The very last line of your response must be exactly of the following format: 'Answer: VALUE' where VALUE is a pure integer. 

Leave a blank line before this final line to completely isolate it from the rest of your explanation.
Ensure there is exactly one space after the colon. Do not include any dollar signs (\$), commas, units, periods, or any other extraneous characters in VALUE. There should be absolutely no additional text, punctuation, or whitespace following the integer.
\end{promptbox}

\begin{promptbox}[title=BBH Object counting]
    Solve the problem.
    
To ensure proper parsing, keep your response as concise as possible.

The very last line of your response must be exactly of the following format: 'Answer: VALUE' where VALUE is an integer.

Do not include any additional text, punctuation, or spaces after the numerical value.
\end{promptbox}

\begin{promptbox}[title=LogiQA]
    You are an expert in logical reasoning. You will answer a logical reasoning question by selecting the correct option from the provided list.

Follow these systematic steps:

1. **Analyze the Query:** Carefully break down the context and the question. Identify the core logical structure, premises, definitions, and specific requirements. Pay special attention to negative constraints (e.g., "NOT", "cannot", "least likely") and the modality required (e.g., "must be true", "most weakens"). Explicitly state the ultimate goal of the question (e.g., "The goal is to find an alternative explanation that directly undermines the causal link").

2. **Formulate Criteria:** Establish clear, precise, and prioritized criteria that the correct answer must satisfy based on your logical analysis. Include negative criteria (what the correct option must *not* do) and distinguish between necessary and sufficient conditions.

3. **Evaluate Options:** Systematically and concisely analyze every option using its zero-based index (e.g., Option 0, Option 1, Option 2, Option 3). Use strong, decisive language to explain exactly how it meets or fails the formulated criteria.

   - For "weaken/strengthen" questions, prioritize options that directly attack or support the core premise or mechanism over those providing indirect, tangential, or correlational impacts.
   
   - For "must be true/inference" questions, ensure the deduction is a necessary consequence of the provided premises. Explicitly reject options that are merely "possible" or require external, unwarranted assumptions.
   
   - Address potential nuances (e.g., correlation vs. causation, ambiguity, shifts in scope), clearly explaining why an incorrect option is logically flawed.
   
4. **Compare and Conclude:** Conduct a rigorous and direct comparative analysis of the most viable options. Be decisive. Explicitly contrast the strengths of the correct option against the specific weaknesses of the runners-up to summarize why the selected option is definitively the best or only correct choice.

5. **Final Answer:** Skip any section heading for this step. On the very last line of your response, provide the final answer in the exact format:

Answer: VALUE
(where VALUE is the single zero-based integer index of the correct option).

CRITICAL FORMATTING RULE: The final line MUST be exactly 'Answer: VALUE' (e.g., 'Answer: 2') with NO additional spaces, NO periods, and NO extra text following the integer. The integer must be the absolute final character of your generated response.

\end{promptbox}

\subsection{CROP Prompts}

\begin{promptbox}[title=GSM8K]
    You will answer a mathematical reasoning question concisely, showing the combined essential calculations leading directly to the final numerical answer.  Omit intermediate labels. Focus on chained calculations. The last line of your response should be of the following format: 'Answer: VALUE' where VALUE is a numerical value.
\end{promptbox}

\begin{promptbox}[title=BBH Object counting]
    You will answer a mathematical reasoning question.
Present your reasoning solely through a single-line numerical calculation, summing only the pure numerical values identified from the question. Do not include item names, descriptions, or separate itemized lists. Do not include any introductory or conversational phrases.
The last line of your response must be of the exact following format: 'Answer: VALUE' where VALUE is a pure numerical integer (without any currency symbols like '\$' or extra characters).
\end{promptbox}

\begin{promptbox}[title=LogiQA]
    You will answer a logical reasoning question. 
    
    Be concise and direct in your reasoning. Avoid unnecessary introductions, restatements of the problem, or verbose explanations. Briefly explain why the correct answer is right and succinctly state why the incorrect options are wrong. 
    
    The very last line of your response MUST be exactly of the following format: 'Answer: VALUE' where VALUE is the zero-based numerical index of the correct answer. Do not add any extra spaces or characters to this final line.
\end{promptbox}

\subsection{Gradient Prompts}

\begin{promptbox}[title=Accuracy Feedback Gradient Prompt]
You are an expert Prompt Engineer. \\
\\
You will give feedback to a variable with the following role: $<$ROLE$>$ response from the language model $<$/ROLE$>$. \\
Here is an evaluation of the variable using a string-based function: \\
\\
Function purpose: The runtime of string-based function that checks if the prediction is correct, where the answer is followed by "Answer :" \\
\\
$<$INPUTS\_TO\_FUNCTION$>$ \{\} \\
\\
\textbf{Ground truth answer(role: correct answer for the query)}: \{\} $<$/INPUTS\_TO\_FUNCTION$>$ \\
\\
$<$OUTPUT\_OF\_FUNCTION$>$ \{\} $<$/OUTPUT\_OF\_FUNCTION$>$ \\
\\
Objective: Your goal is to give feedback and criticism to the variable given the above evaluation output. Our only goal is to improve the above metric, and nothing else. $<$/OBJECTIVE\_FUNCTION$>$ \\
\\
We are interested in giving feedback to the response from the language model for this conversation. Specifically, give feedback to the following span of text: \\
\\
$<$VARIABLE$>$ \{\} $<$/VARIABLE$>$ \\
\\
Given the above history, describe how the response from the language model could be improved to improve the Objective. Be very creative, critical, and intelligent.
\end{promptbox}

\begin{promptbox}[title=Length Regularization Gradient]
You are an expert Prompt Engineer and Efficiency Strategist. \\
\\
You will give feedback to a variable (also known as MODEL\_PROMPT) with the following role: $<$ROLE$>$ structured system prompt to a somewhat capable language model that specifies the behavior $<$/ROLE$>$. \\
\\
$<$MODEL\_PROMPT$>$ \{\} $<$/MODEL\_PROMPT$>$ \\
$<$QUESTION$>$ \{\} $<$/QUESTION$>$ \\
$<$MODEL\_RESPONSE$>$ \{\} $<$/MODEL\_RESPONSE$>$ \\
\\
$<$OBJECTIVE\_FUNCTION$>$Reduce the length of the model response by only few words without hurting the quality, and trying to keep as much reasoning and important stuff and focus on removing the fluff.$<$/OBJECTIVE\_FUNCTION$>$ \\
\\
Large language model takes $<$MODEL\_PROMPT$>$ and $<$QUESTION$>$ as input and generates the $<$MODEL\_RESPONSE$>$. \\
\\
Given the above history, describe how the response from the language model could be updated to achieve the $<$OBJECTIVE\_FUNCTION$>$ without hurting the quality of the response. Be very creative, critical, and intelligent.
\end{promptbox}

\subsection{Optimizer prompt}

\begin{promptbox}[title=System Prompt Update Strategy, colframe=teal!75!black, boxed title style={colback=teal!75!black}]
Here is the role of the variable you will improve: \\
$<$ROLE$>$structured system prompt to a somewhat capable language model that specifies the behavior and strategies for the QA task$<$/ROLE$>$. \\
\\
The variable is the text within the following span: \\
$<$VARIABLE$>$ \{current\_system\_prompt\} $<$/VARIABLE$>$ \\
\\
Here is the context and feedback we got for the variable: \\
\\
$<$CONTEXT$>$ \\
\textit{(The following block repeats for each conversation in the dataset:)} \\
\\
Here is a conversation: \\
\\
$<$CONVERSATION$>$ \\
$<$LM\_SYSTEM\_PROMPT$>$ \{current\_system\_prompt\} $<$/LM\_SYSTEM\_PROMPT$>$ \\
$<$LM\_INPUT$>$ \{questions[i]\} $<$/LM\_INPUT$>$ \\
\\
$<$LM\_OUTPUT$>$ \{forward\_response[i]\} $<$/LM\_OUTPUT$>$ \\
\\
$<$/CONVERSATION$>$ \\
\\
This conversation is potentially part of a larger system. The output is used as response from the language model \\
\\
Here is the feedback we got for structured system prompt to a somewhat capable language model that specifies the behavior and strategies for the QA task in the conversation: \\
\\
$<$FEEDBACK$>$ \\
\{feedback\_response[i]\} \\
$<$/FEEDBACK$>$ \\
\\
$<$REGULARIZATION\_FEEDBACK$>$ \\
\{regularization\_responses[i]\} \\
$<$/REGULARIZATION\_FEEDBACK$>$ \\
\\
$<$/CONTEXT$>$ \\
\\
Improve the variable (structured system prompt to a somewhat capable language model that specifies the behavior and strategies for the QA task) using the feedback provided in $<$FEEDBACK$>$ and $<$REGULARIZATION\_FEEDBACK$>$ tags. \\
Send the improved variable in the following format: \\
\\
$<$IMPROVED\_VARIABLE$>$\{the improved variable\}$<$/IMPROVED\_VARIABLE$>$ \\
\\
Send ONLY the improved variable between the $<$IMPROVED\_VARIABLE$>$ tags, and nothing else.
\end{promptbox}

\end{document}